\documentclass{article}

\usepackage{arxiv}

\usepackage[utf8]{inputenc} 
\usepackage[T1]{fontenc}    
\usepackage[hyphens]{url}   
\usepackage{hyperref}       
\usepackage{booktabs}       
\usepackage{amsfonts}       
\usepackage{nicefrac}       
\usepackage{microtype}      
\usepackage{lipsum}
\usepackage{graphicx}

\usepackage{amssymb,amsmath}
\usepackage{bm}
\usepackage{bbm}
\DeclareMathOperator*{\argmin}{arg\,min}

\usepackage{algorithm}
\usepackage{algorithmic}

\usepackage{xcolor}

\usepackage{tikz}
\usepackage{tikz-qtree}
\usepackage{subcaption}

\usepackage[backend=biber,style=numeric,maxcitenames=99,maxbibnames=99]{biblatex}
\title{Pruning Extensions and Efficiency Trade-Offs for Sustainable Time Series Classification}

\author{
 Raphael Fischer \\
  \textit{XplaiNLP Group}\\
  Johannes Gutenberg University Mainz\\
  Dortmund, Germany\\
  \url{https://orcid.org/0000-0002-1808-5773} \\
   \And
 Angus Dempster \\
  \textit{Department of Data Science \& AI}\\
  Monash University\\
  Melbourne, Australia \\
  \url{https://orcid.org/0000-0002-6809-0482} \\
  \And
 Sebastian Buschjäger \\
  \textit{Lamarr Institute for ML and AI}\\
  TU Dortmund University\\
  Dortmund, Germany \\
  \url{https://orcid.org/0000-0002-2780-3618} \\
  \And
 Matthias Jakobs \\
  \textit{Sleep Medicine Center}\\
  University Medicine Essen\\
  Essen, Germany \\
  \url{https://orcid.org/0000-0003-4607-8957} \\
  \And
 Urav Maniar \\
  \textit{Faculty of Information Technology}\\
  Monash University\\
  Melbourne, Australia \\
  \url{https://orcid.org/0009-0008-1134-9538} \\
  \And
 Geoffrey I. Webb \\
  \textit{Department of Data Science \& AI}\\
  Monash University\\
  Melbourne, Australia \\
  \url{https://orcid.org/0000-0001-9963-5169}
}

\begin{document}
\maketitle
\begin{abstract}
Time series classification (TSC) at scale requires balancing predictive performance against resource consumption, which is subject to the given data, model, and hardware.
This paper introduces a multi-dimensional evaluation framework that allows for explicitly benchmarking and exploring the resource-versus-quality trade-offs in TSC.
To improve efficiency, we present a pruning strategy with bounded error for leading hybrid classifiers---\texttt{Hydra} and \texttt{Quant}---and moreover propose \texttt{Hydrant}, a novel extension that integrates interval representations with random convolutions and dictionary learning.
With over 4000 evaluated configurations across 20 \texttt{MONSTER} datasets, 10 methods, and three compute setups, we systematically analyze how model design, hyperparameters, and hardware choices impact practical TSC performance.
Our results show that pruning can significantly reduce energy consumption by up to 70\% while only marginally lowering predictive quality.
As such, our contributed evaluation framework, methodological extensions, experimental results, and software repository advance TSC toward sustainable and reproducible practice.
\keywords{green AI, sustainability, benchmarking, resource-aware machine learning, deep learning}
\end{abstract}

\section{Introduction}

\begin{figure*}
    \centering
    \includegraphics[width=0.95\linewidth]{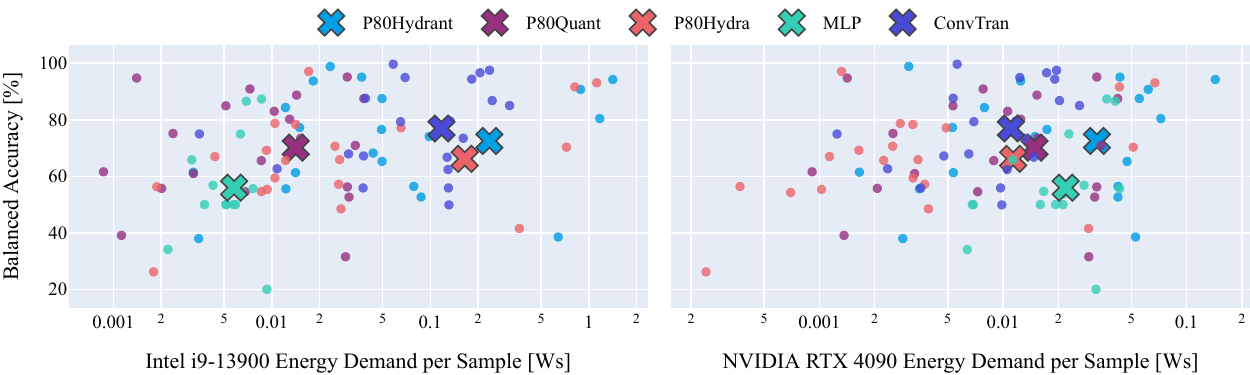}
    \caption{Energy ($x$) versus accuracy ($y$) performance of leading TS classifiers when evaluated for inference efficiency on two processors (left and right). Average performance is indicated by crosses while scatter points represent results on individual datasets, demonstrating intricate performance trade-offs~\cite{fischer_towards_2024}.}
    \label{fig:pareto_performance}
\end{figure*}

Time series (TS) remain essential for representing our modern world in data, with scalability demanding efficient machine learning (ML) methods.
Supervised TS classification (TSC) in particular enables innovative artificial intelligence (AI) applications~\cite{faouzi_time_2024}, such as sleep staging in health~\cite{dietz-terjung_beyond_2025} or land cover mapping in earth observation~\cite{pelletier_temporal_2019,fischer_no_2020}.
Deep learning (DL) allows to train highly accurate models in an end-to-end fashion~\cite{ismail_fawaz_inceptiontime_2020,fawaz_deep_2019,zhao_convolutional_2017,zheng_time_2014}, however traditional ML remains competitive thanks to effective TS feature transformations, such as the random convolutions in \texttt{Hydra}~\cite{dempster_hydra_2023} or interval statistics in \texttt{Quant}~\cite{dempster_quant_2024}.
A recent study not only demonstrated how such hybrids can outperform DL, but also suggests that advancing TSC requires larger volumes of data and more nuanced evaluation frameworks~\cite{dempster_highly_2025}.
Importantly, while many have claimed efficiency improvements for classifying temporal data~\cite{zheng_time_2014,dempster_rocket_2020,cabello_etal_2020_rstsf,dempster_hydra_2023,foumani_improving_2024,dempster_quant_2024,dempster_highly_2025}, we observe a central problem across the literature:

\textbf{TSC experiments are not unified with regard to evaluation configurations and dimensions}, with methods being overly optimized for ``fast and accurate'' performance on small datasets~\cite{dempster_monster_2025}.
Predictive quality and low latency are undeniably important, but do not suffice for understanding the complicated performance trade-offs that hinder sustainable ML development~\cite{fischer_towards_2024}.
As an illustrative example, using a graphics processing unit (GPU) might boost the processing speed but likely consumes more power than only relying on the central processor (CPU).
Unfortunately, TSC works rarely report energy, let alone testing different hardware platforms or investigating the balance between training compute and downstream inference costs.
As an interlude to later experiments, Figure~\ref{fig:pareto_performance} shows how the experimental configuration impacts the performance of leading TSC methods, with regard to energy draw ($x$-axis) and accuracy ($y$-axis).
Each point represents a single evaluation on a \texttt{MONSTER} dataset~\cite{dempster_monster_2025}, with average performance indicated by crosses and \texttt{P80} models denoting our own efficiency-boosted variants of \texttt{Hydra} and \texttt{Quant}.
The strong variance makes it hard to judge overall efficiency and deploying the same classifiers on a GPU completely changes the energy--accuracy performance landscape.

To overcome the questionable ``bigger-is-better paradigm''~\cite{varoquaux_hype_2025} in ML and ensure (environmentally) sustainable progress~\cite{van_wynsberghe_sustainable_2021,fischer_diss,rohde_broadening_2024}, we argue that it is imperative to \textbf{(a) establish a resource-aware evaluation framework for TSC} and \textbf{(b) develop new methods explicitly anchored on scalability and efficiency}.
ML techniques like pruning aim at optimizing trained models for deployment~\cite{kuzmin_pruning_2023,buschjager_joint_2023}, however such possibilities remain under-explored in TSC literature~\cite{uribarri_etal_2024_detach,chen_etal_2024_pocket,salehinejad_etal_2022_srocket} and have not been adapted for frontier methods like \texttt{Hydra}~\cite{dempster_hydra_2023} or \texttt{Quant}~\cite{dempster_quant_2024}.
Inspired by ``green'' methodologies for automated ML~\cite{tornede_towards_2023}, TS forecasting~\cite{fischer_autoxpcr_2024}, and online learning~\cite{kobschall_lift_2025}, our work therefore makes three central contributions for advancing TSC sustainability:

\begin{enumerate}
    \item We introduce a sustainable evaluation framework for benchmarking the efficiency trade-offs of TS classifiers.
    \item We propose a pruning strategy for \texttt{Hydra}~\cite{dempster_hydra_2023} and \texttt{Quant}~\cite{dempster_quant_2024}, as well as a prunable \texttt{Hydrant} extension.
    \item We explore TSC performance across thousands of experiments and unveil configurational impacts on efficiency.
\end{enumerate}

The rest of this paper is structured as follows.
Section~\ref{sec:rw} establishes fundamentals and covers relevant related work.
Section~\ref{sec:meth} sets out formalizations for sustainable TSC benchmarking as well as error-bounded pruning and integration opportunities for frontier TS classifiers.
Section~\ref{sec:exp} presents our experiments, exploring multi-dimensional performance trade-offs between 10 TS classifiers trained across 20 \texttt{MONSTER} datasets~\cite{dempster_monster_2025} and deployed with various configurations (three hardware setups, various pruning rates and batch sizes).
Statistical analyses confirm our three central hypotheses: 
\textbf{TSC inference efficiency is significantly impacted by configuration details (H1), pruning \texttt{Hydra} and \texttt{Quant} reduces energy demand while maintaining high predictive quality (H2), and integrating both methods into \texttt{Hydrant} boosts predictive quality (H3)}.
In Section~\ref{sec:con} we conclude our work, discussing inherent limitations and providing a future work outlook.
Our reproducible codebase, full experimental results, and means for interactive exploration are offered via a software repository at \url{https://github.com/raphischer/efficient-tsc}.
Taken together, our work promotes sustainable development as well as open-science practices in AI and ML, thus providing a unified perspective for advancing TSC while embracing resource-awareness.

\section{Related Work}
\label{sec:rw}

Sensors capturing real-world processes produce large volumes of TS data.
Classification via ML enables various impressive applications~\cite{faouzi_time_2024}, for example deriving satellite land cover maps~\cite{pelletier_temporal_2019,fischer_no_2020} or annotating sleep stages~\cite{dietz-terjung_beyond_2025}.
As such, ML and AI technology enables sustainable innovation, with both examples directly connecting to sustainable development goals (SDG 3 and SDG 13 for health and climate).
However, environmentally sustainable use of ML mandates to also investigate energy efficiency (directly translating to carbon impacts)~\cite{van_wynsberghe_sustainable_2021,rohde_broadening_2024,fischer_diss} and multi-dimensional performance trade-offs along the Pareto front~\cite{varoquaux_hype_2025,fischer_towards_2024}.
While ``green'' methodologies are developed for other ML domains~\cite{tornede_towards_2023,fischer_autoxpcr_2024,kobschall_lift_2025}, we argue that sustainable practice is mostly absent in the TSC domain.

\subsection{Performance, Efficiency, and Datasets}
The treatment of performance in TSC literature remains highly idiosyncratic and largely ignores inference efficiency, with superior results being usually quantified via predictive quality and overall evaluation time (dominated by training) on a single hardware platform~\cite{dempster_monster_2025,dempster_quant_2024,dempster_hydra_2023,dempster_rocket_2020,cabello_etal_2020_rstsf}.
While superficially establishing a level playing field for comparison, this conceals the (potentially overshadowing) inference costs and moreover neglects the variance across experimental configurations---even though practical ML performance is always impacted by choice of hardware, implementation, dataset, and model (alongside hyperparameters)~\cite{fischer_towards_2024,Kriegel_etal_runtime_2017}.

The TSC focus on training connects to the common practice of benchmarking methods on small datasets as for example offered by the \texttt{UCR} archive~\cite{dau_ucr_2019}, which conceal the practical implications of non-linear scaling during inference~\cite{dempster_monster_2025}.
Indeed, historical state-of-the-art TSC methods were burdened by high computational costs that hardly scale beyond toy datasets~\cite{dempster_highly_2025}, while larger archives like \texttt{MONSTER} demand to calibrate the inductive biases in TSC: more data may induce more capable models, but at increased computational cost~\cite{dempster_monster_2025}.
This efficiency trade-off directly connects to the no-free-lunch theorem, which implies that no algorithm is better than any other over all datasets~\cite{wolpert_nfl_1996}.

\subsection{Modeling Approaches}
\label{sec:rw:modeling}
Practitioners today have access to a wide variety of TSC approaches.
For \textbf{traditional ML}, individual TS are commonly transformed into \emph{feature vectors}, which for example encode statistical or frequency domain information~\cite{shekar_selection_2018}. 
Optimization then allows to learn the patterns between TS features and labels, for example via \emph{linear models} (often equipped with parameter regularization), \emph{decision trees} (DTs, encoding hierarchical decision rules), or ensembles comprising multiple weak learners (i.e., random forests consisting of DTs)~\cite{scikit-learn,loning_sktime_2019}.
To mention some famous but somewhat outdated TSC methods, \texttt{DrCIF}~\cite{middlehurst_canonical_2020} constructs a random forest to classify canonical features across different intervals~\cite{lubba_catch22_2019}, \texttt{TS-CHIEF}~\cite{shifaz_ts-chief_2020} uses forests that integrate similarity-based, dictionary-based, and interval-based features, and \texttt{HIVE-COTE} builds a complex ensemble of several TSC classifiers~\cite{middlehurst_hive-cote_2021}.
While most of these approaches hardly scale to larger TS datasets~\cite{dempster_highly_2025}, the more recent \texttt{Quant} achieves high efficiency by classifying simple quantile information over various TS intervals via randomized DTs~\cite{dempster_quant_2024}.

Conversely, the generalizability and GPU scalability of \emph{neural networks} (NNs) popularized to also leverage \textbf{standard DL} for TSC~\cite{loning_sktime_2019}.
Bypassing the explicit feature engineering, it allows for training models as compositions of non-linear transformations in an end-to-end fashion.
Classic multi-layer perceptrons (MLPs) and \emph{fully connected networks} (FCNs) are strong baselines~\cite{wang_time_2017}, while \emph{convolution layers} in CNNs can also catch local patterns over time or channels~\cite{zhao_convolutional_2017}.
As more \textbf{specialized DL} approaches for TSC, \emph{recurrent networks} (RNNs) learn temporal patterns via gated units or long short-term memory (LSTM) cells~\cite{karim_lstm_2018}, \texttt{InceptionTime} combines CNNs with ensembling~\cite{ismail_fawaz_inceptiontime_2020}, and \texttt{ConvTran} incorporates \emph{transformer}-inspired self-attention~\cite{foumani_improving_2024}.
As a hybrid combination of DL and ML, the \texttt{Rocket} variants harness the feature extraction strength of CNNs via so-called random convolution kernels and classify the observed activations via a ridge regressor~\cite{dempster_rocket_2020,dempster_minirocket_2021,tan_multirocket_2022}.
The latest \texttt{Hydra} extension connects this to dictionary learning and symbolic pattern counting, with features describing how often competing kernels within distinct groups are best- and worst-fitting the TS~\cite{dempster_hydra_2023}.
Both \texttt{Quant} and \texttt{Hydra} were extended for scalability across very large datasets~\cite{dempster_highly_2025}, with benchmarking on \texttt{MONSTER} evidencing their superior TSC performance~\cite{dempster_monster_2025} and first experiments suggesting additional accuracy gains from integrating both methods~\cite{maniar2025metalearninggapcombininghydra}.

\subsection{Pruning and Feature Selection}
Overall, it is clear that strong predictive quality in TSC is enabled by complex feature spaces, either via random kernels, interval methods, or latent DL representations, which likely introduce some degree of computational redundancy.
In other learning domains, models are commonly compressed via pruning or quantization strategies, either completely removing individual parameters, neurons, layers, or ensemble members, or projecting them onto a lower-memory encoding~\cite{kuzmin_pruning_2023,buschjager_joint_2023}.
In addition, hardware-focused model optimizations can further increase inference efficiency~\cite{10.1145/3644815.3644967,Buschjaeger/2018a}.

The few TSC works comparable to such techniques focus on \emph{feature selection} during training, for example by ranking interval features via Fisher scores~\cite{cabello_etal_2020_rstsf}.
Some respective approaches were developed for \texttt{Rocket} variants, leveraging evolutionary methods~\cite{salehinejad_etal_2022_srocket}, elastic net regularization~\cite{chen_etal_2024_pocket}, and sequential kernel elimination~\cite{uribarri_etal_2024_detach} for boosting runtime efficiency.
However, these methods cannot be easily applied to the state-of-the-art TSC methods \texttt{Hydra}~\cite{dempster_hydra_2023} and \texttt{Quant}~\cite{dempster_quant_2024}, due to the specific structures of their feature spaces.

\section{Methodology}
\label{sec:meth}

Moving toward a more holistic perspective on TSC performance, efficiency, and sustainability, we now formally introduce our evaluation framework and methodological extensions.

\subsection{Time Series Classification}
\label{sec:meth:tsc}

For any TSC dataset $\mathfrak{D} = \{(\bm{x}_i, y_i)_{1 \leq i \leq N}\}$, let $\mathcal{X}_\mathfrak{D} = \mathbb{R}^{D \cdot L}$ and $\mathcal{Y}_\mathfrak{D} = \{1, 2, \dots, C\}$ denote the data input and target spaces with $C$ class labels.
Each multivariate series $\bm{x}_i= (\bm{x}_{i1}, \bm{x}_{i2}, \dots, \bm{x}_{iD}) \in \mathcal{X}_\mathfrak{D}$ consists of $D$ temporally ordered, $L$-dimensional vectors that are commonly referred to as univariate \emph{channels} with (synchronized) \emph{length} $L$, i.e., $\bm{x}_{ij} = (x_{ij1}, x_{ij2}, \dots, x_{ijL}) \in \mathbb{R}^L$.
$\mathfrak{D}$ comprises $N$ labeled samples from an unknown ground-truth classification function $f^*: \mathcal{X}_\mathfrak{D} \rightarrow \mathcal{Y}_\mathfrak{D}$, and finding a good TS classifier $f_{\bm{\theta}} \approx f^*$ thus corresponds to empirical risk minimization (ERM) on $\mathfrak{D}$, using a loss function $l$ to quantify the fit of the (regularized) model parameters $\bm{\theta}$:

\begin{equation}\label{eq:erm}
    \argmin_{\bm{\theta}} R_\mathfrak{D}(\bm{\theta}) = \argmin_{\bm{\theta}} \frac{1}{N} \sum_{i=1}^N l(f_{\bm{\theta}}(\bm{x}_i), y_i) + \lambda \Omega(\bm{\theta}),
\end{equation}

with $\lambda \Omega(\bm{\theta})$ denoting the regularization term. 
TSC methods differ in their parametrized logic, with feature transformations $\bm{z} = \phi(\bm{x}) \in \mathbb{R}^q$ allowing to reduce the data complexity of $\mathfrak{D}$.
Linear regressors classify TS feature representations via $f_{\bm{\theta}}(\bm{x}) = \bm{\beta}\bm{z} + \bm{b}$, optionally aided by $L_1$ (lasso) or $L_2$ (ridge) regularization within $\Omega(\bm{\theta})$.
DTs instead derive logic rules and classify via $f_{\bm{\theta}}(\bm{x}) = \sum_l p_l \mathbbm{1}[\bm{z} \in R_l]$, indicating whether features satisfy the rules for leaf $l$ with probabilities $p_l$.
Ensembles comprise $M$ base learners and classify via $\bm{f}_{\bm{\theta}}(\bm{x}) = \frac{1}{M} \sum_{m=1}^M f_{m, {\bm{\theta}}} (\bm{z})$, while DL models process latent feature representations across layers of neurons in an end-to-end learning approach.
All classifiers are trained by solving the ERM problem on the annotated data $\mathfrak{D}$ either with a closed-form solution (e.g., for ridge regression~\cite{dempster_quant_2024}) or via stochastic optimization over (batched) TS samples.
Because model predictions $f_{\bm{\theta}}(\bm{x}_i) = \bm{\hat{y}}_i$ might diverge from ground-truth labels, one commonly reports the model quality on hold-out data via metrics such as accuracy or $F_1$ score~\cite{scikit-learn}.
Going beyond predictive capabilities, the performance of classifiers can also be evaluated with respect to resource consumption, for example assessing the running time and energy demand of training or inference~\cite{fischer_towards_2024,fischer_ground-truthing_2025}.

\subsection{Comparing Performance Trade-Offs}
\label{sec:meth:practical_performance}

Benchmarking multi-dimensional performance requires to operationalize the hypothesis that configuration details impact practical performance (H1).
To that end, we build on the sustainable and trustworthy reporting (\texttt{STREP}) methodology~\cite{fischer_towards_2024,fischer_diss} and assess TSC performance via \emph{property functions} $\mu : \mathcal{F} \times \mathcal{C} \rightarrow \mathbb{R}^+$, which map classifiers $f \in \mathcal{F}$ and evaluation configurations $c \in \mathcal{C}$ onto quantifiable metrics like running time, accuracy, or energy.
While omitting the explicit parameter notation $\bm{\theta}$, we consider models $f$ to be fully (hyper-)parametrized and trained, while the configuration $c$ characterizes the given learning task, dataset, software installation, and hardware platform.
To unify properties and compare relative, configuration-specific performance values $\tilde{\bm{\mu}}(f, c)$, \texttt{STREP} introduces \emph{index scaling}:

\begin{equation}\label{eq:index}
    \tilde{\mu}_i(f, c) = \left(\frac{\mu_i(f^*, c)}{\mu_i(f, c)}\right)^{\sigma_i}
    \text{, with } f^* = \argmin_{f'} \mu_i(f', c) \cdot \sigma_i
\end{equation}

Based on the metric-specific direction constant ($\sigma_i \in \{1, -1\}$ for minimization or maximization), this projects ML properties onto the $(0, 1]$ unit scale, where higher values correspond to better performance up to the best empirically observed value $\tilde\mu_i(f^*, c) = 1$.
Moreover, index-scaled properties can be aggregated into a \emph{compound score} that assesses overall performance based on user-controllable weights $\omega_i$~\cite{fischer_towards_2024}:

\begin{equation}\label{eq:compound}
    \mathcal{S}(f, c) = \sum_{i = 1}^p \omega_i \cdot \tilde{\mu}_i(f, c) \text{, with } \sum_{i = 1}^p \omega_i=1 \text{ and } 0 \leq \omega_i \leq 1
\end{equation}

The weighted sum allows to score any model's performance under explicit consideration of use case priorities~\cite{fischer_autoxpcr_2024}, as certain application might for example favor very accurate (safety-critical decisions), fast (real-time applications), or energy-efficient solutions (battery powered edge hardware).

\subsection{Pruning State-of-the-Art Time Series Classifiers}
\label{sec:meth:pruning}

As discussed in Section~\ref{sec:rw:modeling}, \texttt{Hydra}~\cite{dempster_hydra_2023} and \texttt{Quant}~\cite{dempster_quant_2024} are considered state-of-the-art methods for classifying large quantities of TS data~\cite{dempster_monster_2025,dempster_highly_2025}.
The former uses random convolutional kernels with a dictionary learning paradigm~\cite{dempster_hydra_2023}, where features $\bm{z}_{\texttt{Hydra}}$ represent the soft and hard counts of the maximum and minimum responses per kernel group, while classification commences via closed-form ridge regression~\cite{dempster_highly_2025}.
\texttt{Quant} on the other hand calculates features $\bm{z}_{\texttt{Quant}}$ as statistical quantile information on intervals obtained from different TS representations, and classifies via randomized DTs~\cite{dempster_quant_2024}.
Both approaches follow a similar strategy, leveraging complex feature representations that can be subdivided into distinct \emph{feature subsets} (based on kernel groups or intervals), while the classifier learns patterns across this hierarchy.
Focusing on deployment efficiency, we however hypothesize (H2) that not all feature sets are actually informative for an arbitrary but fixed data context, motivating us to apply \textbf{post-training pruning} for retaining only the most relevant features.

\begin{algorithm}[t]
\caption{Pruning \texttt{Hydra} (also adaptable for \texttt{Quant})}\label{alg:pruning}
\begin{algorithmic}
\REQUIRE Dataset $\mathfrak{D}$, prune rate $\zeta$
\ENSURE Pruned feature transform $\Phi_{\texttt{Hydra}}$ and classifier $f$
\STATE Initialize $\phi_{\texttt{Hydra}}$ for $\mathfrak{D}$ \emph{ \# provides indices of $S$ feature sets}
\STATE $\bm{z}'_{\texttt{Hydra}} \gets \phi_{\texttt{Hydra}}(\bm{x})$ \emph{ \# full feature representation}
\STATE $f' \gets \text{ERM on } (\bm{z}'_{\texttt{Hydra}}, \bm{y})$ \emph{ \# intermediate classifier (ridge)}
\STATE $\mathcal{I}_{\texttt{Hydra}} \gets \emptyset$ \emph{ \# importance per feature set (kernel group)}
\FOR {$s \in \{1, \dots, S\}$}
    \STATE $\bar{\beta}_s \gets \frac{1}{|\iota(s)|}\sum_{i \in \iota(s)} |\bm{\beta}_{f', i}|$ \emph{ \#  $\iota(s)$ are the feature indices}
    \STATE $\mathcal{I}_{\texttt{Hydra}} \gets \mathcal{I}_{\texttt{Hydra}} \cup \{(s, \bar{\beta}_s)\}$ \emph{ \# store mean importance}
\ENDFOR
\STATE $r \gets (1 - \zeta) \cdot S$ \emph{ \# number of feature sets to keep} 
\STATE $\pi \gets \operatorname*{argsort}(\mathcal{I}_{\texttt{Hydra}})$ \emph{ \# sort for decreasing importance}
\STATE $\Phi_{\texttt{Hydra}} \gets \{ \phi_{{\texttt{Hydra}}_{\iota(\pi(s))}} \}_{s=1}^{r}$ \emph{ \# keep most relevant}
\STATE $\bm{z}_{\texttt{Hydra}} \gets \Phi_{\texttt{Hydra}}(\bm{x})$ \emph{ \# pruned feature representation}
\STATE $f \gets \text{ERM on } (\bm{z}_{\texttt{Hydra}}, \bm{y})$ \emph{ \# pruned classifier (ridge)}
\RETURN $\Phi_{\texttt{Hydra}}$, $f$
\end{algorithmic}
\end{algorithm}

Our strategy is summarized in Algorithm~\ref{alg:pruning} and follows a two-step approach, using a \emph{pruning rate} $\zeta \in [0, 1]$ that allows to identify a fraction $(1 -\zeta)$ of the most relevant feature sets from an \emph{intermediate classifier} $f'$ trained across all features.
Accounting for the hierarchical nature of both TSC methods, we assess and rank the importance of each feature set $\bar{\beta}_s$ by averaging the associated feature attribution coefficients, which are natively available for many classic ML classifiers~\cite{scikit-learn}.
Finding the optimal ML method for deriving feature importances remains a secondary search problem of its own, however using a linear intermediate model has strong theoretical benefits---it bounds the pruning error (i.e., expected loss in quality) based on a uniform bound $B$ across all features:

\noindent
\textbf{Theorem} \emph{(Bounded pruning error)}:\\
Let $S' \subset \{1,\dots,S\}$ be any subset of the feature sets, representing kernel groups or interval quantiles and defining
\[
\bm{f}_S'(\bm{x}) := \sum_{s \in S'} \sum_{i \in \iota(s)} \bm{\beta}_i \bm{z}_i \text{ , where } \iota(s) \text{ are feature indices}.
\]
Assume all features are upper-bounded by $B$, i.e.:
\[
\|\bm{z}_i\|_\infty := \sup_{\bm{x} \in \mathcal{X}_\mathfrak{D}} |\psi_\xi(\bm{x})_i| \le B
\qquad \text{for all } i \in \{1,\dots,q\} .
\]
Then the uniform approximation error incurred by removing all feature sets not in $S'$ satisfies
\[
\|\bm{f} - \bm{f}_{S'}\|_\infty
\;\le\;
B \sum_{s \notin S'} \sum_{i \in \iota(s)} |\bm{\beta}_i|.
\]

\noindent
\textbf{Proof:}
By definition,
\[
\bm{f}(\bm{x}) - \bm{f}_S'(\bm{x})
= \sum_{s \notin S'} \sum_{i \in \iota(s)} \bm{\beta}_i \psi(\bm{x})_i
= \sum_{s \notin S'} \sum_{i \in \iota(s)} \bm{\beta}_i \bm{z}_i 
\]
Taking absolute values and using the triangle inequality yields
\[
|\bm{f}(\bm{x}) - \bm{f}_{S'}(\bm{x}))|
\le \sum_{g \notin S'} \sum_{i \in \iota(s)} |\bm{\beta}_i|\,|\bm{z}_i|.
\]
Since each kernel is uniformly bounded by $B$, we obtain
\[
|\bm{f}(\bm{x}) - \bm{f}_{S'}(\bm{x})|
\le B \sum_{s \notin S'} \sum_{i \in \iota(s)} |\bm{\beta}_i|.
\]
Taking the supremum over $\bm{x}$ concludes the proof. $\square$

While our proof omits the formal intercept $\bm{b}$ and regularization $\lambda \Omega (\bm{\theta})$ terms of Section~\ref{sec:meth:tsc}, they can be easily included by applying sum and product rules.
Note also that for $\bar{\beta}_s \approx 0$, we obtain $f(x)\approx f_{S \setminus \{s\}}(x)$.
Hence, the straightforward and deterministic pruning strategy of removing all kernels with coefficients close to 0 would only have a minimal impact on performance. 
More formally, assume coefficients are decreasingly sorted via $\pi$ such that
$|\bm{\beta}_{\pi(1)}| \le |\bm{\beta}_{\pi(2)}| \le \dots \le |\bm{\beta}_{\pi(S)}|$,
then retaining only the $r$-largest coefficients ($r = (1-\zeta)\cdot S \ll S$) guarantees
\[
\|f - f_{S'}\|_\infty
\le B \sum_{s=1}^{S-r} \sum_{i \in \iota(s)} |\bm{\beta}_i|.
\]

This lower bound is completely distribution-free and holds uniformly over the entire input space $\mathcal{X}_\mathfrak{D}$, however it should be noted that it assumes a linear sum logic of coefficients.
As such, using another intermediate or final classifier such as DTs for \texttt{Quant} invalidates the bound due to non-linearity in feature logic, however might still perform well in practice. 
While \texttt{Hydra} kernel groups result in equally large feature sets, \texttt{Quant} intervals can have varying amounts of features per set---in this case, $\zeta$ merely approximates the pruning amount, however does not induce a strict relation between $|\psi(\bm{x})|$ and $|\Psi(\bm{x})|$.
Note also that pruning could even act as a regularization that alleviates overfitting, reduces variance, increases bias, and potentially lowers the model generalization error on unseen data.

\subsection{Combining Convolution Kernels and Interval Quantiles}
\label{sec:meth:hydrant}

As \texttt{Hydra} and \texttt{Quant} excel in different data contexts~\cite{dempster_monster_2025}, it is reasonable to assume that a hybrid combination could capture additional patterns and thus achieve higher TSC generalization and accuracy.
Based on this hypothesis (H3), we propose a new \texttt{Hydrant} extension that integrates both feature transformations.
However, the naive concatenation ($\oplus$) of all features---\texttt{Hydra}'s weak pattern counts and \texttt{Quant}'s distributional interval statistics---induces a highly heterogeneous feature space that might be too complex for learning informative patterns.
This challenge can be alleviated via \textbf{asymmetric stacking}, such that \texttt{Hydrant} combines \texttt{Quant} features $\bm{z}_{\texttt{Q}}$ with \texttt{Hydra} logits $\bm{p}_{\texttt{H}}$~\cite{maniar2025metalearninggapcombininghydra}.
The complete strategy is outlined in Algorithm~\ref{alg:hydrant} and can be interpreted as \textbf{meta-learning}, where the obtained logits act as meta-features that inform the final classifier about \texttt{Hydra}'s estimated class likelihood.
$k$-fold cross-validation is applied to mitigate information leakage and overfitting during training, such that multiple \texttt{Hydra} classifiers $f_{\texttt{H}}^{(k)}$ are trained to predict the logits across the full training data.
Collecting the predictions $\bm{p}_{\texttt{H}}^{\text{CV}}$ across all folds and concatenating them with the deterministic \texttt{Quant} features $\bm{z}_{\texttt{Q}}$ yields the complete feature representation $\bm{z}_{\text{meta}}$, which can then be learned by the final classifier $\bm{f}$.
We propose to use randomized DTs, which are already effective for \texttt{Quant}~\cite{dempster_quant_2024} and directly benefit from additional meta-knowledge. During inference it is obtained from the downstream logit predictor $f_{\texttt{H}}$, which was learned on the complete training data but not used for fitting the DTs.
As such, \texttt{Hydrant} corresponds to the \texttt{QFeat-HLogit-ET} approach discussed in prior work~\cite{maniar2025metalearninggapcombininghydra} and classifies TS data via:

\begin{algorithm}[t]
\caption{Training \texttt{Hydrant}}
\label{alg:hydrant}
\begin{algorithmic}
\REQUIRE Dataset $\mathfrak{D}$, folds $K$, prune rate $\zeta$
\ENSURE Logit predictor $f_{\texttt{H}}$, classifier $\bm{f}$ 
\STATE Initialize $\phi_{\texttt{Quant}}$ and $\phi_{\texttt{Hydra}}$ for $\mathfrak{D}$
\STATE $\bm{z}_{\texttt{Q}} \gets \phi_{\texttt{Quant}}(\bm{x})$ \emph{ \# interval-based features}
\STATE $\bm{p}_{\texttt{H}}^{\text{CV}} \gets \mathbf{0}_{N \times C}$ \emph{\# initialize cross-validated \texttt{Hydra} logits}
\FOR{$k \in \{1,\dots,K\}$}
    \STATE Split $\mathfrak{D}$ into training folds $\mathfrak{D}_{-k}$ and validation folds $\mathfrak{D}_{k}$
    \STATE $\bm{z}^{(k)}_{\texttt{H}} \gets \phi_{\texttt{Hydra}}(\bm{x}_{-k})$ \emph{ \# kernel-based features for fold}
    \STATE $f_{\texttt{H}}^{(k)} \gets \text{ERM on } (\bm{z}^{(k)}_{\texttt{H}}, \bm{y}_{-k})$ \emph{ \# train fold ridge classifier}
    \STATE $\bm{p}_{\texttt{H},k}^{CV} = f_{\texttt{H}}^{(k)}(\phi_{\texttt{Hydra}}(\bm{x}_{k}))$ \emph{ \# store predicted fold logits}
\ENDFOR
\STATE $\bm{z}_{\text{meta}} \gets \bm{z}_{\texttt{Q}} \oplus \bm{p}_{\texttt{H}}^{\text{CV}}$ \emph{ \# joint meta-feature representation}
\STATE $\bm{f} \gets \text{ERM on } (\bm{z}_{\text{meta}}, \bm{y})$ \emph{ \# train randomized DTs classifier}
\STATE $f_{\texttt{H}} \gets \text{ERM on } (\phi_{\texttt{Hydra}}(\bm{x}), \bm{y})$ \emph{ \# train final logit predictor}
\IF{$\zeta > 0$}
    \STATE prune $\phi_{\texttt{Quant}}$ and $\bm{f}$ via Algorithm~\ref{alg:pruning}
    \STATE prune $\phi_{\texttt{Hydra}}$ and $f_{\texttt{H}}$ via Algorithm~\ref{alg:pruning}
\ENDIF
\RETURN $f_{\texttt{H}}$, $\bm{f}$
\end{algorithmic}
\end{algorithm}

\begin{equation}\label{eq:hydrant}
\bm{f}(\bm{x}) = \frac{1}{M} \sum_{m=1}^M f_m ( \phi_{\texttt{Quant}}(\bm{x}) \oplus f_{\texttt{H}}(\phi_{\texttt{Hydra}}(\bm{x})))
\end{equation}

The obvious downside of \texttt{Hydrant} lies in the additional computational efforts: $k+1$ \texttt{Hydra} instances need to be fitted during training, from which the final ridge regressor is also run in parallel with \texttt{Quant} for downstream TSC.
However, the overhead efforts during inference can once again be alleviated by pruning, following Algorithm~\ref{alg:pruning} to boost the efficiency of the final logit predictor $f_{\texttt{H}}$ and randomized DTs $\bm{f}$ (the cross-validation \texttt{Hydra} instances are discarded after training).
The overall complexity for classifying $N$ TS of length $L$ with $D$ channels into $C$ classes with any of the presented methods and a pruning rate $\zeta$ is as follows:
For \texttt{Hydra}, the computational complexity is dominated by the convolutional feature transformations~\cite{dempster_hydra_2023}.
Assuming $G_{\texttt{H}}$ kernel groups with $K_{\texttt{H}}$ kernels each, distributed across $d_{\texttt{H}}$ dilation levels, the pruned transformation scales as $\mathcal{O}((1-\zeta) \cdot N \cdot D \cdot L \cdot G_{\texttt{H}} \cdot K_{\texttt{H}} \cdot \log L),$
as the number of dilation levels grows logarithmically with the TS length ($d_{\texttt{H}} \sim \log L$). 
The resulting representation $\phi_{\texttt{Hydra}}(\bm{x}) \in \mathbb{R}^{q}$ comprises counts over the kernels per groups, obtaining dimensionality $q = \mathcal{O}((1-\zeta) \cdot N \cdot G_{\texttt{H}} \cdot \log L).$
Training the (batched) closed-form ridge regression is dominated by the feature matrix and eigendecomposition obtaining $\mathcal{O}(\max(N, q) \cdot \min(N, q)^2)$~\cite{dempster_highly_2025}, while inference scales as $\mathcal{O}(C \cdot q)$ per sample. 
For \texttt{Quant}, the computational complexity is determined by the extraction of quantile features over a fixed set of dyadic intervals across $R_\texttt{Q}$ input representations~\cite{dempster_quant_2024}.
In the worst case, calculating the respective quantiles requires complexity $\mathcal{O}(L \log L)$ per TS (sorting). 
As the number of representations and intervals are constants, this yields pruned feature transformation cost of $\mathcal{O}((1-\zeta) \cdot N \cdot L \log L)$.
The resulting features $\phi_{\texttt{Quant}}(\bm{x}) \in \mathbb{R}^{p}$ scale linearly with the TS length ($p = \mathcal{O}(\zeta \cdot L)$), since the number of quantiles per interval grows proportionally.
Training the randomized DTs under the assumption of balanced trees yields $\mathcal{O}(p \cdot N \log N)$, while memory requirements are dominated by the feature matrix and scale as $\mathcal{O}(N \cdot p) = \mathcal{O}(N \cdot (1-\zeta) \cdot L)$.
For \texttt{Hydrant}, the computational complexity combines the costs of \texttt{Quant} and \texttt{Hydra}, so transforming the features requires $\mathcal{O}((1-\zeta) \cdot N \cdot D \cdot L \cdot G_\texttt{H} \cdot K_\texttt{H} \cdot \log L)$.
The resulting hybrid representation $\bm{z}_{\text{meta}} = \phi_{\texttt{Quant}}(\bm{x}) \oplus \bm{p}_{\texttt{Hydra}}(\bm{x})$ has dimensionality $\mathcal{O}((1-\zeta) \cdot L + C)$.
Training the final randomized decision tree ensemble therefore scales as $\mathcal{O}(((1-\zeta) \cdot L + C) \cdot N \log N)$ assuming balanced trees, while memory requirements are dominated by the feature matrix with complexity $\mathcal{O}(N \cdot ((1-\zeta) \cdot L + C))$.
For another practical consideration, note that \texttt{Hydrant} and all pruned variants can also be implemented with sub-batching for memory-constrained training and inference, as already proposed for the original methods~\cite{dempster_highly_2025}.

\section{Experiments}
\label{sec:exp}

Having formalized our methodology, we now explore practical performance landscapes and trade-offs in TSC.

\subsection{Experimental Setup}
\label{sec:exp:setup}

To investigate TSC efficiency at scale, we benchmark all methods across 20 \texttt{MONSTER} datasets from various domains, spanning between 9K--200K instances ($N$), 2--82 classes ($C$), 1--113 channels ($D$) and lengths between 23--5K ($L$).
All experimental results were averaged over the five pre-defined data splits, further \texttt{MONSTER} details are available in the archive documentation~\cite{dempster_monster_2025}. 
Returning to the methodological separation of Section~\ref{sec:rw}, we tested the default \texttt{sktime} implementations~\cite{loning_sktime_2019} of \texttt{MCDCNN}~\cite{zheng_time_2014}, \texttt{MLP} \& \texttt{ResNet}~\cite{wang_time_2017}, \texttt{FCN}~\cite{zhao_convolutional_2017}, \texttt{LSTMFCN}~\cite{karim_lstm_2018}, and \texttt{InceptionTime}~\cite{ismail_fawaz_inceptiontime_2020}.
Moreover, we evaluated the latest implementations of \texttt{ConvTran}~\cite{foumani_improving_2024} as well as \texttt{Hydra} and \texttt{Quant}~\cite{dempster_highly_2025}, to compare them against our new \texttt{Hydrant} extension and pruned variants (e.g., \texttt{P80\{base\}} denotes a $\zeta = 80\%$ pruning of the base model, with only 20\% retained features).
As we focus on out-of-the-box performance, we used the same hyperparameters and defaults as proposed in the original works and codebases.

To explore hardware effects, we deployed all models on three different execution environments, using two workstations equipped with a) an \texttt{Intel i9-13900} CPU \& \texttt{NVIDIA RTX 4090} GPU (disabled for CPU experiments) and b) an \texttt{Intel i7-6700} CPU.
Quantifying inference performance, we investigated class-balanced accuracy, macro $F_1$ score, and precision for quality, as well as memory size, energy demand, and running time (per sample) for resource demand~\cite{fischer_towards_2024}.
For compound scoring (cf. Equation~\ref{eq:compound}), we prioritize well-balanced performance~\cite{fischer_autoxpcr_2024,fischer_diss} by using equalized weights (i.e., $\omega=1/6 \approx0.167$ for all properties).
Energy was assessed with \texttt{CodeCarbon 3.0.8} and as such only reflects the CPU and GPU demand, not accounting for any overhead estimation errors or embodied impacts~\cite{fischer_ground-truthing_2025}.
Following the calls for sustainable reporting~\cite{fischer_towards_2024}, we estimate the total amount of energy consumed by our experiments to $762+145=907$ kWh (representing the dominating development efforts as well as final runs), translating to roughly $\text{345 kg CO}_\texttt{2}\text{eq}$, assuming a German energy grid emission intensity of $0.38\text{ kg CO}_\texttt{2}\text{eq} / \text{kWh}$.
Training was performed with batches of 32 instances, while inference was tested with batch sizes between $2^4$ and $2^{14}$.
Unless otherwise specified, the results report the model performance with the optimal (lowest energy) batch size.
While the following focuses on crucial takeaways, we provide all benchmarking results along with the extended paper version and codebase at \url{https://github.com/raphischer/efficient-tsc}, allowing readers to also interactively explore TSC performance trade-offs with the \texttt{STREP} dashboard~\cite{fischer_towards_2024}.

\subsection{Experimental Results}
\label{sec:exp:results}

\begin{figure*}[t]
    \centering
    \includegraphics[width=0.95\linewidth]{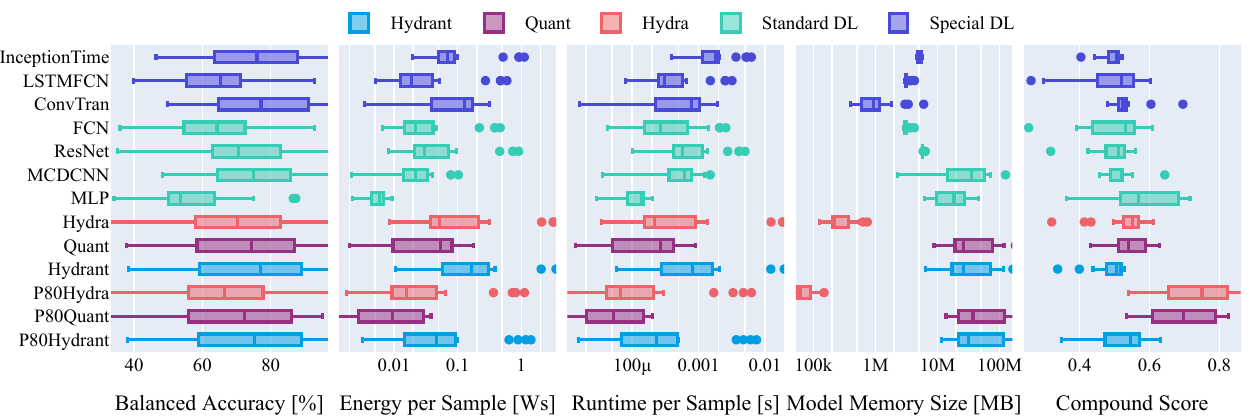}
    \caption{Inference performance trade-offs among TSC models ($y$-axis) deployed across datasets (box plots) on the \texttt{Intel i9-13900} CPU.
    Accuracy (first plot) comes at the price of resources (second to fourth), while compound scores (fifth) reveal the overall balance across all evaluation dimensions.}
    \label{fig:model_stats}
\end{figure*}

Recalling the opener Figure~\ref{fig:pareto_performance}, TSC models can be generally observed to trade resources (energy demand per sample on $x$-axis) against quality (balanced accuracy on $y$-axis), with observable deviations across the tested datasets (scatter variance) and execution environment (differences between left and right plot) that connect to no-free-lunch~\cite{wolpert_nfl_1996}.
Digging deeper into the performance results on the \texttt{i9} CPU (left side), we visualize the performance of all evaluated models across all datasets along different performance dimensions via box plots in Figure~\ref{fig:model_stats}.
While most models have roughly similar accuracy distributions (first plot), we see bigger differences and trade-offs for energy demand (second), running time (third), and memory (fourth).
For example, \texttt{ConvTran} and \texttt{Hydrant} have the highest energy consumption but obtain the best prediction performance, whereas \texttt{Quant} and \texttt{Hydra} are much faster and energy-saving.
The compound score aggregated from index-scaled properties (right plot, cf. Equation~\ref{eq:index} and Equation~\ref{eq:compound}) assesses the quality--resource trade-off across all properties, where our pruned variants of \texttt{Hydra} and \texttt{Quant} score best.
Note that our score equally weights all evaluation dimensions, however the higher prediction quality of \texttt{Hydrant} and specialized DL approaches at the cost of more energy might be justified for safety-critical applications---the adequate balance of performance dimensions remains subject to use case priorities~\cite{fischer_towards_2024,fischer_autoxpcr_2024}.

\begin{figure*}[t]
    \centering
    \includegraphics[width=0.95\linewidth]{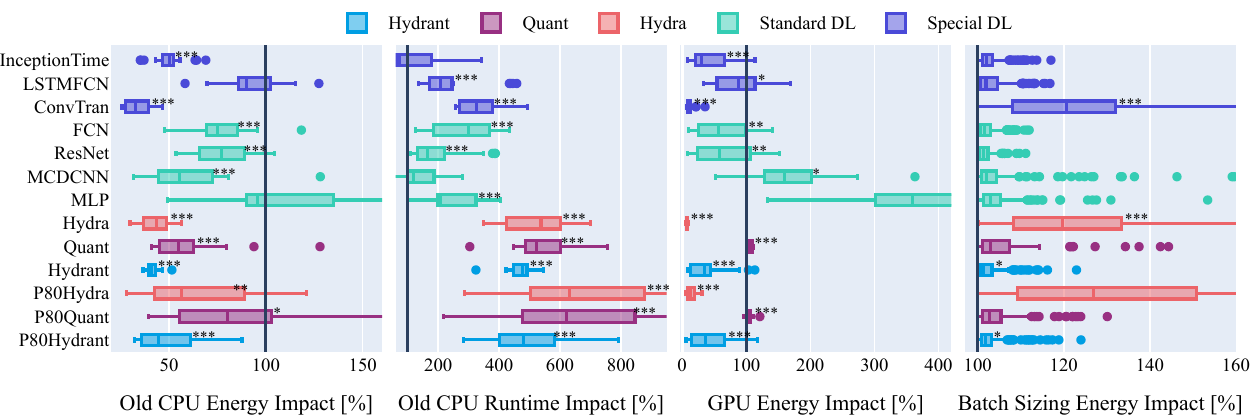}
    \caption{Impacts on model performances (box plots), compared against the \texttt{i9} configuration. Using the older \texttt{i7} CPU lowers energy demand (first plot) but increases running time (second), while GPU deployment (third) and batch sizing (fourth) can also significantly influence any model's energy efficiency.}
    \label{fig:config_impact}
\end{figure*}

Figure~\ref{fig:config_impact} shows how different configurations impact the relative efficiency, using the \texttt{i9} CPU with optimal batch size as reference (vertical bars at $x=100\%$).
The left plot demonstrates how switching to the older \texttt{i7} platform generally lowers the energy demand per sample during inference---however, this also comes with drastically increased running time (up to a factor of ten) for all models but \texttt{InceptionTime} and \texttt{MCDCNN} (second plot).
These energy-versus-runtime trade-offs vary, as some models like \texttt{MLP} and \texttt{LSTMFCN} behave rather similarly (slightly lower energy at higher running time), while larger differences can for example be observed for \texttt{ConvTran} or \texttt{Hydrant}.
The third plot evidences the heavily reduced energy demand of GPU inference, with strong gains for the specialized DL approaches and \texttt{Hydra} variants (less than 10\% of the CPU energy).
\texttt{Quant} is not optimized for GPU usage and thus behaves nearly the same as on the CPU, whereas the \texttt{MLP} and \texttt{MCDCNN} models have higher energy demand (up to 400\%).
On the right, we see the impact of using different batch sizes instead of the most efficient one (different for each model, dataset, and environment).
While often only accounting for a few percent of increased energy, we see that the \texttt{ConvTran} and \texttt{Hydra} classifiers should be carefully tuned---for them, arbitrarily chosen batch sizes often result in a 20--30\% higher energy demand.
Next to all boxes, statistically significant distribution effects are indicated by asterisks (*/**/*** for $p<0.05/0.01/0.001$, per Wilcoxon signed-rank tests for columns 1-3 and a Friedman test across batch sizes in column 4, details can be obtained from our \texttt{run\_analysis.py} script).
As such, the results support our configuration impact hypothesis (H1) for nearly all assessments and demonstrate the limitations of comparing TSC performance on a singular fixed hardware setup.

\begin{figure*}[t]
    \centering
    \includegraphics[width=0.95\linewidth]{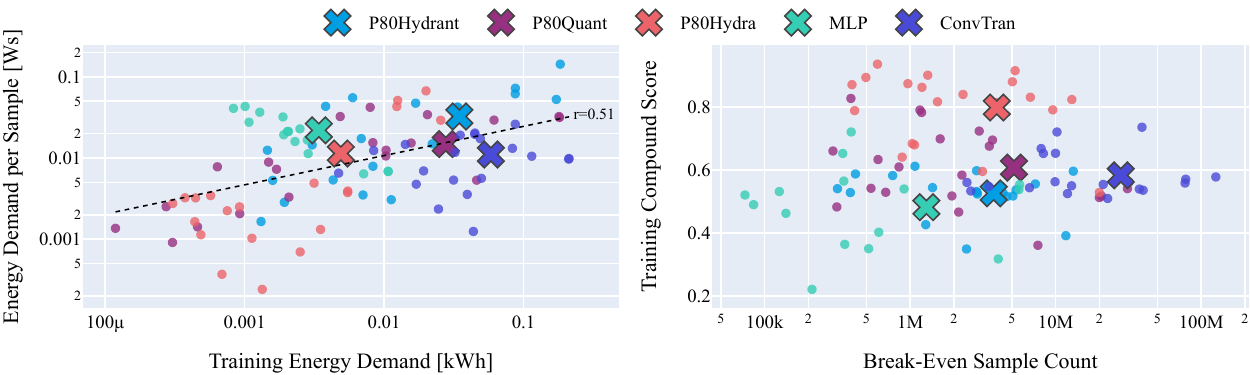}
    \caption{Training ($x$) and inference energy demand ($y$) vary heavily but generally correlate across all datasets on the GPU (left). Our pruned variants have solid overall compound scores (right $y$) and score mediocre in inference amortization ($x$, samples processed for equaling consumed training energy).}
    \label{fig:train_trades}
\end{figure*}

Whereas TSC literature mostly reports overall evaluation time, Figure~\ref{fig:train_trades} reveals additional trade-offs between the training and inference stage.
As in Figure~\ref{fig:pareto_performance}, the scatter points represent individual evaluations and averaged performance (crosses) for all the datasets using the \texttt{RTX 4090} GPU.
On the left, we observe a mediocre correlation between training and inference energy demand, with strong deviations from the fitted regression curve demonstrating that reporting on evaluation times alone hardly suffices for understanding operational energy costs.
On the right, the $x$-axis denotes the number of processable samples until reaching the break-even point with training energy, where our pruned variants are located between the low-accuracy \texttt{MLP} and highly accurate \texttt{ConvTran}. 
However, when also taking predictive performance into account, pruning \texttt{Quant} and \texttt{Hydra} results in better overall compound scores ($y$-axis).
Compared to the score in Figure~\ref{fig:model_stats}, we here include properties for training running time and energy demand, balancing them by assessing regular accuracy and weighted $F_1$ (i.e., combining a total of ten properties with $\omega=0.1$).

\begin{figure*}[t]
    \centering
    \hfill
    \includegraphics[width=0.3\linewidth]{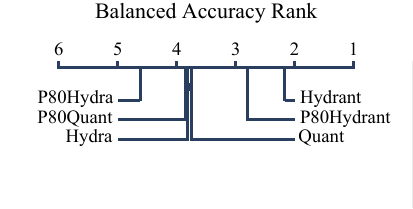}
    \hfill
    \includegraphics[width=0.3\linewidth]{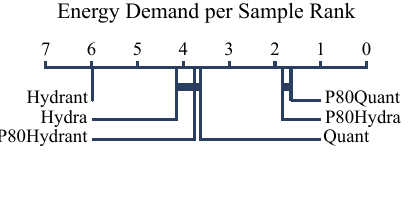}
    \hfill
    \includegraphics[width=0.3\linewidth]{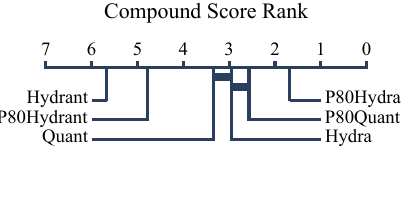}
    \hfill
    \caption{Critical differences~\cite{demsar_statistical_2006} among original and pruned TSC models, showing the significant energy efficiency gains of pruning (higher ranks in middle plot) as well as boosted accuracy of \texttt{Hydrant} (left). Best overall performance is observed for pruned \texttt{Hydra} and \texttt{Quant} (right).}
    \label{fig:cd}
\end{figure*}

To better understand the implications of pruning and integrating \texttt{Hydra} and \texttt{Quant}, Figure~\ref{fig:cd} explores the critical differences~\cite{demsar_statistical_2006} in performance improvements across all configurations (i.e., datasets and environments), with statistically significant results indicated by missing horizontal bars.
When ranking the original model against their drastically pruned variants, a significant drop in accuracy rank (left plot) can be observed for \texttt{Hydra} and \texttt{Hydrant}.
However, the pruned classifiers also achieve significantly better ranks regarding energy demand (middle) and compound score (right), thus confirming our pruning hypothesis (H2).
\texttt{Hydrant} and its pruned variant can moreover be seen to achieve significantly higher quality than the base models (left), at the cost of significantly more energy (middle), resulting in low compound score ranks (right)---nevertheless, this confirms our \texttt{Hydrant} quality improvement hypothesis (H3).

\begin{figure*}[t]
    \centering
    \includegraphics[width=0.95\linewidth]{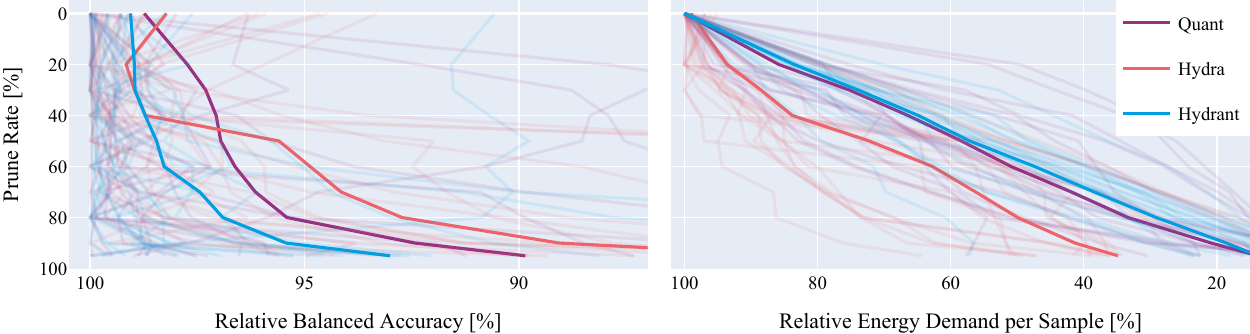}
    \caption{Pruning rate ablation study on model quality (left) and linearly decreasing energy demand (right), averaged across all datasets (thin lines).}
    \label{fig:pruning_ablation}
\end{figure*}

Lastly, while so far having used a pragmatic pruning rate of $\zeta=80\%$, we can also investigate the respective convergence.
Figure~\ref{fig:pruning_ablation} explores implications of pruning with $\zeta \in [0, 0.95]$ for the \texttt{i9} CPU, with thick lines representing averages and thin lines indicating ablation results for individual datasets.
The left plot demonstrates how less than 10\% of the highest accuracy is lost on average for $\zeta<90\%$, with models occasionally even gaining some accuracy at lower rates but slowly deteriorating when $\zeta>40\%$.
On the right, the energy demand can be observed to linearly drop with pruning, with \texttt{Hydra} receiving slightly lower energy savings (while also being more strongly impacted in quality).
Pruning with $\zeta=80\%$ preserves an average relative quality of 92--97\% while decreasing the resource demand by 50--70\%, hence we fixed this rate for all other experiments.
Once again, the results empirically confirm that pruning drastically reduces the energy demand of hybrid TSC methods while maintaining or occasionally even increasing predictive quality.

\section{Conclusion}
\label{sec:con}

Our study has presented a nuanced perspective on evaluating and boosting TSC efficiency, thus paving a way to harness ML potentials for scalable and sustainable TS mining.
This directly addresses the limitations of existing TSC works, particularly regarding their questionable evaluation scope and efficiency claims.
Our methodological extensions connect to the potentials discussed by others~\cite{maniar2025metalearninggapcombininghydra,uribarri_etal_2024_detach,chen_etal_2024_pocket,salehinejad_etal_2022_srocket} and directly advance the efficiency and accuracy of the frontier TSC methods \texttt{Hydra}~\cite{dempster_hydra_2023} and \texttt{Quant}~\cite{dempster_quant_2024,dempster_highly_2025}.
Our benchmark experiments suggest that no-free-lunch also holds for TSC~\cite{wolpert_nfl_1996}---\texttt{Quant} stands out in CPU evaluations, GPUs boost the deployment efficiency of \texttt{Hydra} and specialized DL~\cite{foumani_improving_2024,ismail_fawaz_inceptiontime_2020}, and the latter excel in quality at cost of energy.
Importantly, this demonstrates that the ``optimal'' model performance remains subject to use case requirements and configurations (H1), however feature pruning (H2) and integration of feature representations (H3) resulted in significant performance boosts.
While our study remains somewhat limited in the amount of tested configurations, we offer our codebase to extend the analysis to other datasets, models, and hardware setups.
While here we focus on environmental sustainability, the social and economic dimensions (e.g., related to explainability or fairness) remain for future work~\cite{rohde_broadening_2024,fischer_diss}.
The same goes for hardware-aware optimization~\cite{10.1145/3644815.3644967,Buschjaeger/2018a} or automating the search for efficient TS classifiers, possible under consideration of use case priorities~\cite{fischer_autoxpcr_2024}, thus representing a research opportunity for green AutoML~\cite{tornede_towards_2023}.
For any such future work, our findings evidence that sustainable progress in the field requires researchers to perform rigorous and holistic evaluations that truly investigate the multi-dimensional and interlocked performance spaces in TSC.

\printbibliography

\end{document}